\pdfoutput=1

\documentclass[11pt]{article}

\usepackage{ACL2023}

\usepackage{times}
\usepackage{latexsym}

\usepackage{arydshln}
\usepackage{multirow}
\usepackage{multibib}
\usepackage{booktabs}

\usepackage{amsfonts}
\usepackage{times}
\usepackage{latexsym}
\usepackage{color}
\usepackage{tikz}
\usepackage{pgfplots}
\usepackage{graphicx}
\usepackage{subfig}
\usepackage{caption}
\usepackage{amsmath}
\usepackage{makecell}
\usepackage{hyperref}
\usepackage{amssymb}

\definecolor{myblue}{RGB}{24,116,205}
\definecolor{myred}{RGB}{178,34,34}
\definecolor{ugreen}{RGB}{124,205,124}
\definecolor{ublue}{RGB}{39,64,139}
\definecolor{myyellow}{RGB}{227,207,87}
\usepackage[T1]{fontenc}

\usepackage[utf8]{inputenc}

\usepackage{microtype}
\usepackage{inconsolata}
\renewcommand{\thefootnote}{\fnsymbol{footnote}}
\title{Bridging the Granularity Gap for Acoustic Modeling}

\author{Chen Xu\textsuperscript{1}, Yuhao Zhang\textsuperscript{1}, Chengbo Jiao\textsuperscript{2}, Xiaoqian Liu\textsuperscript{1}, Chi Hu\textsuperscript{1}, \\
 {\bf Xin Zeng\textsuperscript{1}}, 
{\bf Tong Xiao\textsuperscript{1,3}\footnotemark[1] , Anxiang Ma\textsuperscript{1,3}, Huizhen Wang\textsuperscript{1,3}, Jingbo Zhu\textsuperscript{1,3}} \\
  \textsuperscript{1}School of Computer Science and Engineering, Northeastern University, Shenyang, China\\
  \textsuperscript{2}Beijing University of Posts and Telecommunications \\
  \textsuperscript{3}NiuTrans Research, Shenyang, China \\
  \texttt{\{xuchennlp,yoohaozhang,liuxiaoqian0319\}@outlook.com} \\
  \texttt{\{xiaotong,maanxiang,wanghuizhen,zhujingbo\}@mail.neu.edu.cn} 
  }

\begin{document}
\maketitle
\begin{abstract}
\footnotetext[1]{Corresponding author.}
\renewcommand{\thefootnote}{\arabic{footnote}}
While Transformer has become the de-facto standard for speech, modeling upon the fine-grained frame-level features remains an open challenge of capturing long-distance dependencies and distributing the attention weights.
We propose \textit{Progressive Down-Sampling} (PDS) which gradually compresses the acoustic features into coarser-grained units containing more complete semantic information, like text-level representation.
In addition, we develop a representation fusion method to alleviate information loss that occurs inevitably during high compression.
In this way, we compress the acoustic features into 1/32 of the initial length while achieving better or comparable performances on the speech recognition task.
And as a bonus, it yields inference speedups ranging from 1.20$\times$ to 1.47$\times$.
By reducing the modeling burden, we also achieve competitive results when training on the more challenging speech translation task
\footnote{The code is available at {\href{https://github.com/xuchennlp/S2T}{https://github.com/xuchennlp/\\S2T.}}}.


\end{abstract}

\section{Introduction}

Despite the success in speech processing tasks like automatic speech recognition (ASR) \cite{Lu_Interspeech2020, Zhang_Corr2021} and speech translation (ST) \cite{Anastasopoulos_iwslt2022}, how to encode the speech features effectively is an open problem.
Different from modeling based on discrete tokens in natural language processing, a standard paradigm for acoustic encoding is taking the continuous features extracted by signal processing as input.

\begin{figure}[tbp]
  \centering
    \begin{tikzpicture}
    \footnotesize{
      \begin{axis}[
        ymajorgrids,
        xmajorgrids,
        grid style=dashed,
        width=.5\textwidth,
        height=0.2\textwidth,
        legend columns=1,
        legend style={
          draw=none,
          line width=1pt,
        },
        legend style={at={(0.73,1.0)}, anchor=north,
        nodes={scale=0.8, transform shape}},
        xmin=10, xmax=60,
        ymin=0, ymax=10,
        xtick={20, 30, 40, 50},
        ytick={0, 4, 8},
        xlabel=\footnotesize{\footnotesize Length ratio},
        ylabel=\footnotesize{\footnotesize Percentage (\%)},
        ylabel style={yshift=-1.5em},
        xlabel style={yshift=0.0em},
        scaled ticks=false,
        yticklabel style={/pgf/number format/precision=2,/pgf/number format/fixed},
        ]
        \addplot[teal, line width=1pt,mark=*,mark repeat=2,mark size=1pt] 
        file {data/librispeech_length_ratio2.txt};
    \end{axis}
  }
  \end{tikzpicture}
  \caption{The distribution of the length ratio between the speech features (frame-level) and corresponding transcriptions (subword-level) on Librispeech test-clean set.}
  \label{length_ratio}
\end{figure}
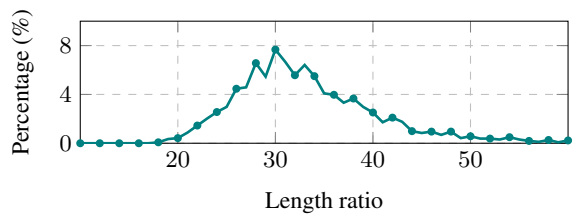

In this process, the framing operation generates a lengthy sequence consisting of fine-grained features.
For example, frame-level feature sequences are dozens of times longer than the corresponding subword-level transcriptions (see Figure \ref{length_ratio}).
Such an input leads to the difficulties of capturing long-distance dependencies and distributing the attention weights among semantically incomplete modeling units \cite{Han_Interspeech2019}.
\citet{Rfou_AAAI2019} also demonstrates that the fine-grained character-level modeling yield significantly inferior performance compared with coarse-grained word-level counterparts.
In addition, the long sequence also results in prohibitive computation costs due to the quadratic complexity of self-attention.

A popular method is to compress fine-grained features to form coarser-grained modeling units by stacking multiple down-sampling modules before encoding \cite{Dong_ICASSP2018, Berard_IEEE2018}.
Unfortunately, it does not work well when we increase the down-sampling (DS) ratio like 16 and 32 to attain semantically more complete units.
This is in line with the intuition that it is difficult to compress dozens of frames directly into a single unit \cite{SAYOOD20181}.
It is something like that a few principal components can not preserve all the information in the principal component analysis method \cite{wold1987pca}.

We analyze the compression process in the stack method to shed light on the reason for the failure.
We find that the similarity of representations in adjacent positions degrades after down-sampling, which leads to the non-trivial issue of information loss and increases the difficulty of subsequent compression.
To address this issue, we propose \textit{Progressive Down-Sampling} (PDS), which gradually aggregates the input frame-level representations into semantically more complete text-level representations, like character-level or subword-level units.
In this way, the model behaves as a natural language processing (NLP) system, which distributes the attention only among a short sequence and enables reinvesting saved computational resources into a deeper model or a bigger batch.
Nevertheless, we find that information loss still occurs as the compression ratio increases.
As a remedy, we combine the coarse-grained high-level representation at top layers and fine-grained low-level representations at bottom layers that may preserve information lost during compression.



Our method outperforms the standard Transformer and Conformer on the ASR and ST tasks, and has the potential for application on a wide variety of acoustic modeling tasks.
It also makes a trade-off between performance and computational speedup.
With an extremely high compression ratio of 32, our method achieves better or comparable performances on the ASR task while yielding inference speedups ranging from 1.20$\times$ to 1.47$\times$.
The lower ratio of 8 or 16 brings remarkable improvement and relatively modest speedup.
On the more challenging ST task, PDS also helps convergence and shows competitive results.


\section{Related Work}

Unlike text that has explicit boundaries, audio is in general represented as continuous signals.
Although researchers have explored straightforward modeling based on the raw audio signal \cite{Steffen_Interspeech2019}, the popular method for segmentation is framing with a frame size of 25ms and a frame shift of 10ms \cite{oppenheim_1999}.
This short frame shift allows the continuity of the speech signal, and the overlapping segments help to avoid information loss between consecutive frames.

However, the fine-grained frame-level features may not be suitable for the state-of-the-art Transformer architectures \cite{Vaswani_nips2017}.
The lengthy sequences composed of semantically incomplete units lead to the difficulties of capturing long-distance dependencies and distributing the attention weights to the most related positions.
Researchers \cite{Salesky_ACL2019, Salesky_ACL2020} investigate phoneme-level methods. 
For example, one can average frame-level features within phoneme-like units.
But this needs a non-trivial recognizer for phoneme alignment.

Motivated by the work in the efficient models \cite{Beltagy_CoRR2020}, researchers alleviate the modeling difficulty by the improved self-attention mechanisms \cite{Han_Interspeech2019, Alastruey_Corr2021, Papi_Corr2021}.
However, they ignore the inherent problem of long sequence modeling and the cross-attention module still suffers from the same issue.

Another line of research is to down-sample the fine-grained features for shorter sequences \cite{Chan_Corr2015, Bahdanau_ICASSP2016}.
A popular approach is to pass the features through a stack of strided convolutional layers before encoding \cite{Dong_ICASSP2018, Berard_IEEE2018}.
But the stack method does not work well in practice due to the loss of information due to consecutive convolutional operations. 
As a way to address this, several research groups use the progressive method to down-sample the acoustic sequence \cite{Peddinti_IEEE2018, Huang_Interspeech2020, Han_Interspeech2020, Burchi_Corr2021}.
The key difference is that the previous studies motivate efficient computation and only explore a modest compression ratio of 8.
Recently, \citet{Andrei22Uconv} reduce the lengths by 16 times at the intermediate layers.
Our work aims to develop a model that takes audio as input but behaves as an NLP style to ease the modeling burden, which requires extremely high compression.
Although there is no special design for efficiency, our method still yields a remarkable speedup.

Another related open problem for acoustic encoding is the variable information caused by silence or noise.
Researchers develop adaptive selection \cite{Zhang_EMNLP2020} or dynamic down-sampling methods \cite{Na_APSIPA2019, Zhang_Interspeech2019} for avoiding useless features.
However, the granularity of the filtered representation is still far from ideal.
These two methods are complementary and we will explore their combination in future work.

\usetikzlibrary{shapes.geometric}
\usetikzlibrary{fit,backgrounds}
\definecolor{ugreen}{RGB}{124,205,124}
\definecolor{myblue}{RGB}{24,116,205}
\definecolor{myred}{RGB}{178,34,34}
\definecolor{ublue}{RGB}{39,64,139}
\definecolor{myyellow}{RGB}{227,207,87}
\usetikzlibrary{shadows}
\newlength{\heightSpan}
\begin{figure*}[tbh!]
  \centering
  \begin{tikzpicture}[scale=0.9]
    \tikzstyle{every node}=[scale=0.9]
    \setlength{\heightSpan}{1.2em}
    \tikzstyle{encoderStyle} = [minimum width=7em,inner sep=4pt,rounded corners=1pt,draw,fill=ugreen!50]
    \tikzstyle{downModuleStyle} = [minimum width=7em,inner sep=4pt,rounded corners=1pt,draw,fill=myblue!30,shape=trapezium,trapezium stretches,trapezium angle=85]
    \tikzstyle{representStyle} = [minimum height=1em,rounded corners=1pt,draw,fill=myyellow!25,minimum width=1em]
    \tikzstyle{convStyle} = [minimum width=6em,inner sep=2pt,rounded corners=1pt,draw,fill=orange!25]
    \tikzstyle{layerStyle} = [minimum width=6em,inner sep=2pt,rounded corners=1pt,draw,fill=teal!25]
    \node[] (speech1) at (0,0) {\small Speech feature};
    \node[downModuleStyle] (down1) at ([yshift=1.1em]speech1.north) {\small Down-sampling};
    \node[] (dot1) at ([yshift=0.6em]down1.north) {......};
    \node[downModuleStyle] (down2) at ([yshift=1em]dot1.north) {\small Down-sampling};
    \node[encoderStyle] (encoder1) at ([yshift=1.2em]down2.north) {\small Encoder layer};
    \node[] (dot2) at ([yshift=0.6em]encoder1.north) {......};
    \node[encoderStyle] (encoder2) at ([yshift=1em]dot2.north) { \small Encoder layer};
    \draw [very thick,decorate,decoration={brace},ugreen] ([xshift=-0.5em,yshift=0.2em]encoder1.south west) to node [midway,name=final] {} ([xshift=-0.5em,yshift=-0.2em]encoder2.north west);
    \draw [very thick,decorate,decoration={brace},ublue!60] ([xshift=-3.2em,yshift=0.2em]down1.south west) to ([xshift=-3.2em,yshift=-0.2em]down2.north west);
    \node[] (t1) at ([xshift=-1.8em,yshift=1.8em]encoder1.west) {\small $L \times$};
    \node[] (t2) at ([xshift=-1.8em,yshift=1.8em]down1.west) {\small  $N \times$};

    \draw[->] ([yshift=-0.2em]speech1.north) -- (down1.south);
    \draw[->] (down1.north) -- ([yshift=0.2em]dot1.south);
    \draw[->] ([yshift=-0.2em]dot1.north) -- (down2.south);
    \draw[->] (down2.north) -- (encoder1.south);
    \draw[->] (encoder1.north) -- ([yshift=0.2em]dot2.south);
    \draw[->] ([yshift=-0.2em]dot2.north) -- (encoder2.south);
    \draw[->] (encoder2.north) -- ([yshift=2.3em]encoder2.south);

    \node[] (a) at ([yshift=-1em]speech1.south) {(a) Stack};

    \draw[-,very thick,dotted,ublue] ([xshift=2.8em,yshift=4.3em]encoder1.east) -- ([xshift=2.8em,yshift=-8.3em]encoder1.east);
    

    \node[] (speech21) at ([xshift=14em]speech1.east) {\small Speech feature};
    \node[downModuleStyle] (down21) at ([yshift=2.0em]speech21.north) {\small Down-sampling};
    \node[encoderStyle] (encoder21) at ([yshift=1.6em]down21.north) {\small Encoder layer};
    \node[] (dot21) at ([yshift=1em]encoder21.north) {......};
    \node[encoderStyle] (encoder22) at ([yshift=1.4em]dot21.north) {\small Encoder layer};
    \draw [very thick,decorate,decoration={brace},ugreen] ([xshift=-0.5em,yshift=0.3em]encoder21.south west) to node [midway,name=final] {} ([xshift=-0.5em,yshift=-0.3em]encoder22.north west);

    \draw[->] ([yshift=-0.3em]speech21.north) -- (down21.south);
    \draw[->] (down21.north) -- (encoder21.south);
    \draw[->] (encoder21.north) -- (dot21.south);
    \draw[->] ([yshift=-0.1em]dot21.north) -- (encoder22.south);
    \draw[->] (encoder22.north) -- ([yshift=3.3em]encoder22.south);

    \node[] (text) at ([xshift=9em]speech21.east) {\small Down-sampling};
    \node[representStyle,minimum width=7em] (rep1) at ([yshift=1.0em]text.north) {};
    \node[convStyle] (conv1) at ([yshift=0.7*\heightSpan]rep1.north) {\small Conv1D};
    \node[layerStyle] (layer1) at ([yshift=0.75*\heightSpan]conv1.north) {\small Layer Norm};
    \node[circle,minimum size=1em,draw,fill=orange!35] (cir) at ([yshift=0.7*\heightSpan]layer1.north) {};
    \node[representStyle,minimum width=4em] (rep2) at ([yshift=0.7*\heightSpan]cir.north) {};
    \node[] (tmp) at ([yshift=0.4em]speech21.north) {};
    \node[] (tmp2) at ([yshift=0.8em]encoder22.north) {};
    \node[] (block) at ([xshift=-5.4em,yshift=0.4em]encoder22.north) {\textcolor{ublue}{\small \ Stage $ \ m$}};

    \draw[->] (rep1.north) -- (conv1.south);
    \draw[->] (conv1.north) -- (layer1.south);
    \draw[->] (layer1.north) -- (cir.south);
    \draw[->] (cir.north) -- (rep2.south);

    \draw[-] (cir.west) -- (cir.east);
    \draw[-] (cir.north) -- (cir.south);
    \node[circle,minimum size=1em,draw] (pe1)  at ([xshift=2em]cir.east) {};
    \draw[-] (pe1.west) .. controls ([xshift=0.6em,yshift=0.6em]pe1.west) and ([xshift=-0.6em,yshift=-0.6em]pe1.east) .. (pe1.east);
    \draw[->] (pe1.west) -- (cir.east);

    \node[] (t3) at ([xshift=-1.9em,yshift=1.5em]encoder21.west) {\small $L_m \times$};
    \node[] (t4) at ([xshift=-1.3em]down21.west) {\textcolor{black}{$1 \times$}};
    \node[] (tmp3) at (t3) {};

    \begin{pgfonlayer}{background}

      \node [rectangle,rounded corners=2pt,dashed,line width=1.5pt,minimum width=12em,minimum height=9.5em,draw=ublue!80,drop shadow,fill=white] [fit = (encoder21) (tmp) (tmp2) (tmp3)] (box1) {};

      \node [rectangle,rounded corners=2pt,line width=1pt,minimum width=10em,minimum height=7em,draw=black!75,drop shadow,fill=white,dashed] [fit = (rep1) (rep2)] (box2) {};

    \end{pgfonlayer}

    \draw[-,black!75,line width=1pt,dashed] (down21.east) -- ([yshift=0em]box2.south west);
    \draw[-,black!75,line width=1pt,dashed] (down21.east) -- ([yshift=0em]box2.north west);
    \node[] (t5) at ([xshift=-1.3em]box1.west) {\textcolor{ublue}{\small $M \times$}};

    \node[] (a) at ([yshift=-1em,xshift=6em]speech21.south) {(b) PIA};
    \node[representStyle,minimum width=1em,minimum height=0.6em] (rep3) at ([yshift=2.2em,xshift=1.2em]box2.north west) {};
    \node[circle,minimum size=0.5em,draw] (pe2)  at ([yshift=-0.8em]rep3.south) {};
    \draw[-] (pe2.west) .. controls ([xshift=0.6em,yshift=0.6em]pe2.west) and ([xshift=-0.6em,yshift=-0.6em]pe2.east) .. (pe2.east);

    \node[anchor=west] (tuzhu1) at ([xshift=0.5em]rep3.east) {\small Representation};
    \node[anchor=west] (tuzhu2) at ([xshift=0.5em]pe2.east) {\small Position Embedding};
    \node[] (tt1) at ([xshift=3em]pe2) {};
    \node[] (tt2) at (tuzhu2) {};
    \node[] (tt3) at (tuzhu1) {};

    \begin{pgfonlayer}{background}
      \node [rectangle,rounded corners=2pt,minimum width=10em,minimum height=2em,draw=black!70,dashed,line width=1pt] [fit = (tt1) (tt2) (tt3)] (box3) {};
    \end{pgfonlayer}


  \end{tikzpicture}
  \caption{Comparison of the Stack and PDS methods.}
  \label{comparison}
\end{figure*}
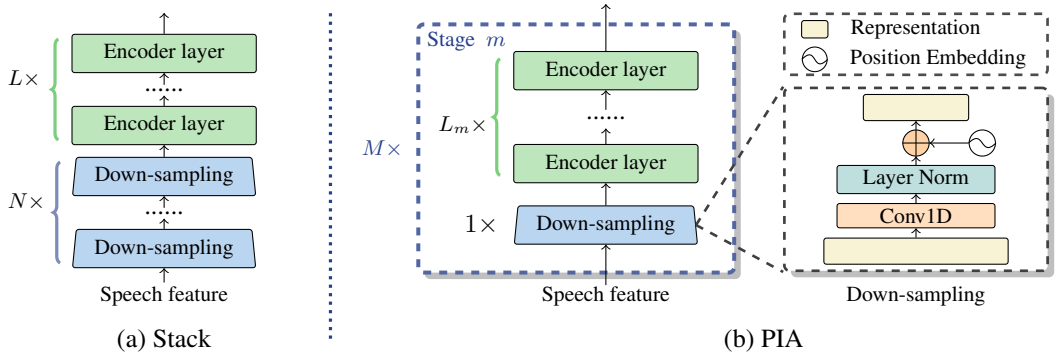

\section{The Method}

\subsection{Why Is Information Lost?}
\label{why}

Down-sampling increases the granularity of modeling units while reducing the sequence length by aggregating the adjacent features.
Following previous work \cite{Dong_ICASSP2018, Berard_IEEE2018}, input speech features are fed into a stack of 2 convolutions with a stride of 2, followed by a number of encoder layers (see Figure \ref{comparison} (a)).
For generating text-level representations, it is natural to stack more down-sampling layers for a high down-sampling ratio.
But it fails in our preliminary experiments (see Table \ref{why_fusion}).

This motivates us to investigate the changes in representation during down-sampling.
We define the \textit{similarity} of representation as the average cosine similarity of each unit to the surrounding units within a small window \cite{Kim22Squeezeformer}.
High similarity indicates that the representation is easier for compression.

\begin{figure}[tbp]
    \begin{tikzpicture}
    \footnotesize{
      \begin{axis}[
        ymajorgrids,
        xmajorgrids,
        grid style=dashed,
        width=.28\textwidth,
        height=0.28\textwidth,
        legend columns=3,
        legend entries={Win-1, Win-2, Win-3},
        legend style={
          draw=none,
          line width=0.8pt,
        },
        ytick={20,40,60,80,100},
        legend style={at={(1.15,1.17)}, anchor=north,
        nodes={scale=0.8, transform shape}},
        xmin=0, xmax=4,
        ymin=0, ymax=100,
        xlabel=\footnotesize{Number of DS},
        ylabel=\footnotesize{Similarity (\%)},
        ylabel style={yshift=-2em},
        xlabel style={yshift=0.0em},
        scaled ticks=false,
        yticklabel style={/pgf/number format/precision=2,/pgf/number format/fixed},
        ]
        \addplot[teal, line width=0.8pt,mark=*,mark size=1pt]
        coordinates{
            (0, 82.53) (1, 61.59) (2, 36.82) (3, 25.92) (4, 19.20)
        };
        \addplot[orange, line width=0.8pt,mark=star,mark size=1.7pt] 
        coordinates{
            (0, 77.30) (1, 53.85) (2, 29.87) (3, 19.35) (4, 12.07)
        };
        \addplot[blue!65, line width=0.8pt,mark=square,mark size=1pt] 
        coordinates{
            (0, 72.56) (1, 48.66) (2, 26.03) (3, 15.66) (4, 8.71)
        };
    \end{axis}
  }
\hspace{3.4cm}
  \footnotesize{
    \begin{axis}[
      ymajorgrids,
      xmajorgrids,
      grid style=dashed,
      width=.28\textwidth,
      height=0.28\textwidth,
      xmin=0, xmax=12,
      ymin=0, ymax=100,
      xtick={0,4,8,12},
      ytick={20,40,60,80,100},
      xlabel=\footnotesize{Layer},
      ylabel style={yshift=-2em},
      xlabel style={yshift=0.0em},
      scaled ticks=false,
      axis y line*=right,
      yticklabel style={/pgf/number format/precision=2,/pgf/number format/fixed},
      ]
      \addplot[teal, line width=0.8pt,mark=*,mark size=1pt] 
      file {data/layer_cos_sim_win1.txt};
      \addplot[orange, line width=0.8pt,mark=star,mark size=1.7pt] 
      file {data/layer_cos_sim_win2.txt};
      \addplot[blue!65, line width=0.8pt,mark=square,mark size=1pt] 
      file {data/layer_cos_sim_win3.txt};  
  \end{axis}
  
}
\draw[-] (0,0) --(0,2.9);
  \end{tikzpicture}
  \caption{Left: the similarity after each down-sampling. Right: the similarity at each Layer in a standard Transformer-based ASR model. Win-$d$ represents the window size of $d$. 
  }
  \label{cos_sim}
\end{figure}
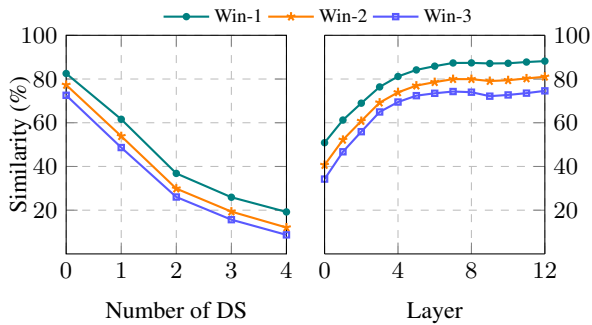

Firstly, we train a Transformer-based \cite{Vaswani_nips2017} ASR model with a down-sampling ratio of 16 by 4 stacked compression operations on the 960h LibriSpeech dataset, then show the similarity on the test-clean set.
As shown in Figure \ref{cos_sim} (Left), the input speech features have an extremely high similarity due to the overlapping framing.
However, the similarity degrades drastically after each down-sampling.
This indicates that the subsequent down-sampling processes are difficult to compress the dissimilar representation while fully preserving the information.
We refer to this issue as information loss caused by stacked down-sampling.

Now a new question arises: how to increase the representation similarity and alleviate the information loss?
An intuitive conjecture is that the context modeling increases the similarity due to the strong preference for short-distance dependency \cite{Sperber_Interspeech2018, Xu_ACL2021}.
Figure \ref{cos_sim} (Right) shows the similarity at each layer of the encoder in a standard Transformer with a down-sampling ratio of 4.
As we expected, the similarity gradually increases from bottom to top.
This inspires us to develop a progressive method that encodes context information sufficiently before each down-sampling.

\subsection{Progressive Down-Sampling}

We propose a \textit{Progressive Down-Sampling} (PDS) method to compress the fine-grained frame-level features into semantically more complete units, like text-level representations.
See Figure \ref{comparison} (b) for an overview of PDS.

The encoding is divided into representation down-sampling and context interaction processes.
Given the input speech features $H_0$, a down-sampling module compresses it by a single 1-D convolution layer.
To address varying lengths, position encoding is introduced into the normalized representation.
Following the finding in Section \ref{why}, we use several encoder layers to capture the contextual dependencies for high similarity.

We define each run of down-sampling and interaction processes as a \textit{stage}.
The model runs for $M$ stages and gradually obtains coarser-grained representations $\{H_1, H_2, \cdots, H_M\}$.
Note that the stack method can be seen as a specific case of PDS: it consists of two stages, where the first stage does not include the context interaction process.


Our method also draws inspirations from the field of computer vision \cite{He_CVPR2016, Wang_Corr2021} and NLP \cite{Dai_NIPS2020}.
While they employ a similar design concept, i.e., gradually reducing the sequence length of the representation, it remains an open problem how to compress the fine-grained and lengthy speech features into text-level representations that are easier for modeling.

\subsection{Representation Fusion}
\renewcommand{\thefootnote}{\arabic{footnote}}
As the inherent nature of compression, information loss still occurs inevitably, especially as the down-sampling ratio increases.
Motivated by previous methods to make full use of the multi-level representations \cite{Wang_ACL2019}, one way to further alleviate this problem is to fuse the coarse-grained representation and the finer-grained representations \cite{Zhao_CVPR2017, Zhang_ECCV2020, Dai_NIPS2020}.
Then the final output representation $H^{o}$ can be defined as:
\begin{eqnarray}
    H^o = \mathcal{F} (H_1, \cdots, H_{M})
\end{eqnarray}
where $\mathcal{F}(\cdot)$ is the fusion function\footnote{We omit the input feature because it is extracted by signal processing rather than encoding by the model.}.

Now the key challenge is how to combine the representations with different lengths.
The premise is to align them to the length of the coarsest-grained representation at the top of the encoder $H_M$.
We resort to multiple simple but effective convolution modules to transform the finer-grained representations outputted in the bottom stages to the shape of $H_M$.
Concretely, the stride and kernel size of convolution for the representation $H_m$ are set to the $L_{H_m}/L_{H_M}$, where $L_{H}$ represents the sequence length of the representation $H$.

\definecolor{myyellow}{RGB}{255,215,0}
\definecolor{ugreen}{RGB}{124,205,124}
\definecolor{ublue}{RGB}{39,64,139}
\usetikzlibrary{shapes.geometric,fit,backgrounds,shadows}
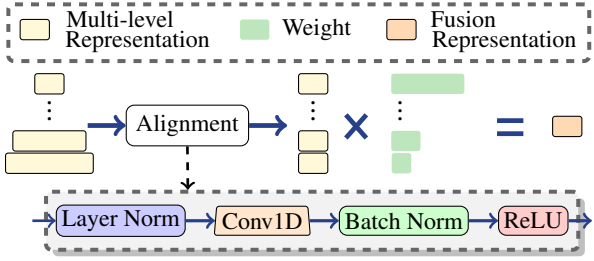
\begin{figure}[t!]
\centering
\begin{tikzpicture}
    \tikzstyle{representStyle} = [minimum height=0.7em,rounded corners=1pt,draw,fill=myyellow!15,minimum width=1em]
    \tikzstyle{prob} = [minimum height=0.7em,rounded corners=1pt,fill=ugreen!50,minimum width=1em]

    \node[representStyle,minimum width=3em] (rep1) at (0,0) {};
    \node[representStyle,minimum width=2.5em] (rep2) at ([yshift=0.4em,xshift=0em]rep1.north) {};
    \node[](dot1) at ([yshift=0.4em,xshift=0]rep2.north) {.};
    \node[](dot2) at ([yshift=-0.05cm]dot1.north) {.};
    \node[](dot3) at ([yshift=-0.05cm]dot2.north) {.};
    \node[representStyle,minimum width=1em] (rep3) at ([yshift=1.6em,xshift=0em]rep2.north) {};

    \node[draw,rounded corners=3pt] (alig) at ([yshift=1.3em,xshift=3.2em]rep1.east) {\small Alignment};

    \node[representStyle,minimum width=1em] (rep21) at ([yshift=0em,xshift=7.5em]rep1.east) {};
    \node[representStyle,minimum width=1em] (rep22) at ([yshift=0.4em]rep21.north) {};
    \node[](dot21) at ([yshift=0.4em,xshift=0em]rep22.north) {.};
    \node[](dot22) at ([yshift=-0.05cm]dot21.north) {.};
    \node[](dot23) at ([yshift=-0.05cm]dot22.north) {.};
    \node[representStyle,minimum width=1em] (rep23) at ([yshift=1.6em]rep22.north) {};
    
    \draw[-,line width=1.8pt,ublue] ([xshift=0.6em,yshift=0.9em]rep22.east) -- ([xshift=1.3em,yshift=0.1em]rep22.east);
    \draw[-,line width=1.8pt,ublue] ([xshift=0.6em,yshift=0.1em]rep22.east) -- ([xshift=1.3em,yshift=0.9em]rep22.east);

    \draw[->,line width=1.8pt,ublue] ([xshift=-1.3em]alig.west) -- (alig.west);
    \draw[->,line width=1.8pt,ublue] (alig.east) -- ([xshift=1.3em]alig.east);

    \node[prob,minimum width=0.5em] (rep31) at ([yshift=0em,xshift=2.5em]rep21.east) {};
    \node[prob,minimum width=1em] (rep32) at ([yshift=0.4em,xshift=0.15em]rep31.north) {};
    \node[](dot31) at ([yshift=0.4em,xshift=-0.2em]rep32.north) {.};
    \node[](dot32) at ([yshift=-0.05cm]dot31.north) {.};
    \node[](dot33) at ([yshift=-0.05cm]dot32.north) {.};
    \node[prob,minimum width=2.5em] (rep33) at ([yshift=1.6em,xshift=0.75em]rep32.north) {};

    \node[representStyle,minimum width=1em,fill=orange!30] (rep42) at ([yshift=0.5em,xshift=5em]rep32.east) {};

    \draw[-,line width=1.8pt,ublue] ([xshift=2.6em,yshift=0.7em]rep32.east) -- ([xshift=3.5em,yshift=0.7em]rep32.east);
    \draw[-,line width=1.8pt,ublue] ([xshift=2.6em,yshift=0.25em]rep32.east) -- ([xshift=3.5em,yshift=0.25em]rep32.east);

    \node[minimum height=1em,inner sep=2pt,rounded corners=3pt,draw,fill=blue!20,anchor=west] (layerNorm) at ([xshift=-4.5em,yshift=-4em]alig.north) {\small Layer Norm};
    \node[minimum height=1em,inner sep=2pt,rounded corners=1pt,draw,fill=orange!20,shape=trapezium,trapezium stretches,trapezium angle=85,anchor=west] (conv) at ([xshift=1em]layerNorm.east) {\small Conv1D};
    \node[minimum height=1em,inner sep=2pt,rounded corners=3pt,draw,fill=green!20,anchor=west] (batchNorm) at ([xshift=1em]conv.east) {\small Batch Norm};
    \node[minimum height=1em,inner sep=2pt,rounded corners=3pt,draw,fill=red!20,anchor=west] (relu) at ([xshift=1em]batchNorm.east) {\small ReLU};

    \draw[->,ublue,line width=1pt]([xshift=-0.8em]layerNorm.west)--(layerNorm.west);
    \draw[->,ublue,line width=1.2pt](layerNorm.east)--(conv.west);
    \draw[->,ublue,line width=1.2pt](conv.east)--(batchNorm.west);
    \draw[->,ublue,line width=1.2pt](batchNorm.east)--(relu.west);
    \draw[->,ublue,line width=1.2pt](relu.east)--([xshift=0.8em]relu.east);

    \begin{pgfonlayer}{background}
      \node [rectangle,rounded corners=2pt,dashed,line width=1.5pt,minimum width=18em,minimum height=2em,draw=black!60,drop shadow,fill=gray!10] [fit = (layerNorm) (relu)] (box1) {};
      \draw[->,line width=1pt,dashed] (alig.south) -- ([xshift=-4.25em]box1.north);
    \end{pgfonlayer}
  

  \node[representStyle,minimum width=1em] (tuzhu1) at ([yshift=1.5em,xshift=-0.5em]rep3.north) {};
  \node[] (tuzhu11) at ([xshift=2.4em,yshift=0.35em]tuzhu1.east) {\footnotesize Multi-level};
  \node[] (tuzhu12) at ([xshift=3em,yshift=-0.4em]tuzhu1.east) {\footnotesize Representation};

  \node[prob,minimum width=1em] (tuzhu2) at ([yshift=0em,xshift=7em]tuzhu1.east) {};
  \node[] (tuzhu22) at ([xshift=1.6em]tuzhu2.east) {\footnotesize Weight};

  \node[representStyle,minimum width=1em,fill=orange!30] (tuzhu3) at ([yshift=0em,xshift=4.5em]tuzhu2.east) {};
  \node[] (tuzhu33) at ([xshift=1.6em,yshift=0.35em]tuzhu3.east) {\footnotesize Fusion};
  \node[] (tuzhu333) at ([xshift=3em,yshift=-0.4em]tuzhu3.east) {\footnotesize Representation};

  \node[](tmp)at ([xshift=0.3em]tuzhu1.west) {};
  \node[](tmp2)at ([xshift=-0.6em]tuzhu333.east) {};
  \node[](tmp3)at ([yshift=-0.8em]tuzhu11.north) {};
  \node[](tmp4)at ([yshift=0.8em]tuzhu12.south) {};

  \begin{pgfonlayer}{background}
    \node [rectangle,rounded corners=2pt,dashed,line width=1.5pt,minimum width=10em,draw=black!60,fill=white] [fit = (tmp) (tmp2) (tmp3) (tmp4)] (box1) {};
  \end{pgfonlayer}

  \end{tikzpicture}
\caption{The representation fusion method. It aligns multi-level representations to the same shapes then combines them.}
\label{alignment}
\end{figure}

Drawing on the design of the convolution module in Conformer, the representation fusion method is shown in Figure \ref{alignment}.
After alignment operation $\mathcal{A}(\cdot)$, we employ a simple linear combination for fusion:
\begin{eqnarray}
\mathcal{F} (H_1, \cdots, H_{M}) = \sum_{m=1}^M W_k \cdot \textrm{LN}(\mathcal{A}(H_m))
\end{eqnarray}
where $\textrm{LN}(\cdot)$ is the layer normalization function. 
$W_m \in \mathbb{R}$ is a learnable scalar to weight the aligned representations.
The weights are initialized to the same values, then learned together with other parameters during training.

\subsection{PDS Settings}
\label{details}

\begin{table}[t]
  \centering
  \footnotesize
  \begin{tabular}{l|l|l}
    \toprule
    Setting & Stride & Layer \\
    \midrule
    Stack-4 & 2-2 & 0-12 \\
    PDS-Base-8 & 2-2-1-2 & 3-3-3-3 \\
    PDS-Base-16 & 2-2-2-2 & 2-2-6-2 \\
    PDS-Base-32 & 2-2-2-2-2 & 2-2-3-3-2 \\
    \midrule
    Stack-4 & 2-2 & 0-30 \\
    PDS-Deep-8 & 2-2-1-2 & 7-7-7-9 \\
    PDS-Deep-16 & 2-2-2-2 & 5-5-12-8 \\
    PDS-Deep-32~~~ & 2-2-2-2-2~~~ & 5-5-7-7-6~~~ \\
    \bottomrule
  \end{tabular}
  \caption{Settings of PDS. "Stack-4" represents the standard method. "PDS-Base-$R$" and "PDS-Deep-$R$" denote an encoder of 12 layers and 30 layers with a down-sampling ratio of $R$, respectively. "Stride" and "Layer" separated by "-" represent the stride of the down-sampling module and the number of layers in each stage from bottom to top.}
  \label{PDS_setting}
\end{table}

In this work, we construct 8 settings under different encoder depths and down-sampling ratios (see Table \ref{PDS_setting}) with the following design philosophy\footnote{
The sophisticated down-sampling designs, fusion approaches, and optimized settings could potentially perform better, but these are beyond the scope of this paper. We briefly opt for the simple design and expect to bring insights for future studies.
}:

\begin{table*}[tbh!]
\centering
\resizebox{\textwidth}{!}{
\footnotesize
  \begin{tabular}{c|l|c|c|c|c|r|c|c|c|c|l|c}
    \toprule
    \multirow{2}*{Group} & \multirow{2}*{Setting} & \multirow{2}*{L} & \multirow{2}*{$d_h$} & \multirow{2}*{$d_{ff}$} & \multirow{2}*{$h$} & \multirow{2}*{{\footnotesize \#Params}} & \multicolumn{2}{c|}{{\footnotesize dev}} & \multicolumn{2}{c|}{{\footnotesize test}} & \makecell[c]{\multirow{2}*{Avg.}} & \multirow{2}*{{\footnotesize Speedup}} \\
    \cmidrule(lr){8-9} \cmidrule(lr) {10-11}
    & & & & & & & {\footnotesize clean} & {\footnotesize other} & {\footnotesize clean} & {\footnotesize other} & & \\
    \midrule
    \multicolumn{13}{l}{Transformer} \\
    \midrule
    \multirow{5}*{(A)} & Stack-4$^*$ & \multirow{5}*{12} & \multirow{5}*{256} & \multirow{5}*{2048} & \multirow{5}*{4} & 30M & 3.80 & 8.90 & 4.40 & 9.00 & - & - \\
    & Stack-4 & & & & & 30M & 3.88 & 9.26 & 4.49 & 9.42 & 6.80 & 1.00$\times$ \\
    & PDS-Base-8  & & & & & 30M & 3.57 & 8.63 & 3.85 & 8.58 & \textbf{6.11 (-0.69)} & 0.99$\times$ \\
    & PDS-Base-16 & & & & & 30M & 3.71 & 8.73 & 3.74 & 9.02 & 6.26 (-0.54) & 1.14$\times$ \\
    & PDS-Base-32 & & & & & 31M & 4.13 & 9.31 & 4.21 & 9.31 & 6.69 (-0.11) & 1.20$\times$ \\
    \specialrule{0em}{1pt}{1pt}
    \cdashline{1-13}
    \specialrule{0em}{1pt}{1pt}
    \multirow{5}*{(B)} & Stack-4$^*$ & \multirow{5}*{12} & \multirow{5}*{512} & \multirow{5}*{2048} & \multirow{5}*{8} & 71M & 3.20 & 8.00 & 3.40 & 7.90 & - & - \\
    & Stack-4 & & & & & 71M & 3.53 & 8.15 & 3.67 & 7.96 & 5.78 & 1.00$\times$ \\
    & PDS-Base-8  & & & & & 75M & 3.17 & 7.46 & 3.47 & 7.47 &\textbf{5.35 (-0.43)} & 1.08$\times$ \\
    & PDS-Base-16 & & & & & 76M & 3.34 & 7.73 & 3.37 & 7.85 &5.53 (-0.25)& 1.34$\times$ \\
    & PDS-Base-32 & & & & & 82M & 3.32 & 7.94 & 3.64 & 7.85 & 5.65 (-0.13) & 1.47$\times$ \\
    \specialrule{0em}{1pt}{1pt}
    \cdashline{1-13}
    \specialrule{0em}{1pt}{1pt}
    \multirow{4}*{(C)} & Stack-4 & \multirow{4}*{30} & \multirow{4}*{256} & \multirow{4}*{2048} & \multirow{4}*{4} & 53M & 3.80 & 8.51 & 4.33 & 8.61 & 6.25 & 1.00$\times$ \\
    & PDS-Deep-8  & & & & & 53M & 3.34 & 7.90 & 3.50 & 7.79 & 5.59 (-0.66) & 1.03$\times$ \\
    & PDS-Deep-16 & & & & & 54M & 3.15 & 7.83 & 3.38 & 7.79 & \textbf{5.50 (-0.75)} & 1.19$\times$ \\
    & PDS-Deep-32 & & & & & 55M & 3.26 & 7.77 & 3.33 & 7.88 & 5.52 (-0.73) & 1.27$\times$  \\
    \midrule
    \multicolumn{11}{l}{Conformer} \\
    \midrule
    \multirow{5}*{(D)}& Stack-4$^\dag$ & \multirow{5}*{12} & \multirow{5}*{256} & \multirow{5}*{2048} & \multirow{5}*{4} & 50M & - & - & 3.05 & 8.36 & - & - \\
    & Stack-4     & & & & & 45M & 3.02 & 7.23 & 3.21 & 7.29 & 5.15 & 1.00$\times$ \\
    & PDS-Base-8  & & & & & 46M & 2.87 & 7.07 & 3.01 & 7.20 & 4.98 (-0.17) & 0.97$\times$ \\
    & PDS-Base-16 & & & & & 46M & 2.93 & 6.97 & 2.99 & 7.03 &\textbf{4.93 (-0.22)} & 1.14$\times$ \\
    & PDS-Base-32 & & & & & 47M & 2.95 & 7.15 & 3.04 & 7.12 & 5.02 (-0.13) & 1.20$\times$ \\
    \specialrule{0em}{1pt}{1pt}
    \cdashline{1-13}
    \specialrule{0em}{1pt}{1pt}
    \multirow{5}*{(E)} & Stack-4$^\ddag$ & \multirow{5}*{12} & \multirow{5}*{512} & \multirow{5}*{2048} & \multirow{5}*{8} 
                        & 109M & 2.90 & 6.60 & 3.00 & 6.70 & - & - \\
    & Stack-4     & & & & & 113M & 2.77 & 6.47 & 2.82 & 6.72 & 4.66         & 1.00$\times$ \\
    & PDS-Base-8  & & & & & 113M & 2.72 & 6.54 & 3.03 & 6.38 & 4.65 (-0.01) & 0.97$\times$ \\
    & PDS-Base-16 & & & & & 114M & 2.73 & 6.31 & 2.72 & 6.40 & \textbf{4.52 (-0.14)} & 1.23$\times$ \\
    & PDS-Base-32 & & & & & 119M & 2.70 & 6.67 & 2.95 & 6.81 & 4.75 (+0.09) & 1.25$\times$ \\
    \bottomrule
  \end{tabular}
}
\caption[Caption for LOF]{WER on the 960h LibriSpeech ASR corpus. L: the number of encoder layers. $d_h$: the hidden dimension. $d_{ff}$: the feed-forward dimension. $h$: the number of attention heads. \#Params: the number of parameters. 
The speedup is computed during inference on the test-clean set with a beam size of 5 and batch size of 100k (except 50k for a bigger Conformer due to the GPU limitation). $*$, $\dag$, and $\ddag$ stand for the results reported in fairseq\protect\footnotemark , wenet\protect\footnotemark , and espnet\protect\footnotemark ~respectively.}
\label{res_asr}
\end{table*}

\begin{itemize}
    \item 
    During down-sampling, a bigger window size involves more context information while increasing the difficulty of down-sampling due to the lower similarity.
    Referring to framing, we use an empirical setting of kernel size $=5$ and stride $=2$.
    \item 
    Different from the design in the field of computer vision \cite{He_CVPR2016}, we keep the same hidden dimensions in the whole encoding process.
    The detailed comparisons and analyses are shown in Appendix \ref{appendix_compare_dim}.
    \item 
    We allocate fewer layers at the bottom stage for efficient computation due to the longer sequence. The major computations are concentrated in the intermediate stages. This leads to sufficient encoding and computation acceleration, as shown in Table \ref{layers}.
\end{itemize}


\section{Experiments}

We evaluate our method on the LibriSpeech and AISHELL-1 ASR datasets, and the MuST-C En-De ST dataset. Details about the data and model settings are described in Appendix \ref{appendix_exp}.

\subsection{Results of ASR}
\label{results_asr}

\footnotetext[4]{\href{https://github.com/pytorch/fairseq/blob/main/examples/speech_to_text/docs/librispeech_example.md}{https://github.com/pytorch/fairseq/blob/main/examples/\\speech\_to\_text/docs/librispeech\_example.md}}
\footnotetext[5]{\href{https://github.com/wenet-e2e/wenet/blob/main/examples/librispeech/s0/README.md}{https://github.com/wenet-e2e/wenet/blob/main/example\\s/librispeech/s0/README.md}}
\footnotetext[6]{\href{https://github.com/espnet/espnet/blob/master/egs/librispeech/asr1/RESULTS.md}{https://github.com/espnet/espnet/blob/master/egs/\\librispeech/asr1/RESULTS.md}}

\textbf{LibriSpeech} Table \ref{res_asr} shows the results on the 960h LibriSpeech corpus.
We compare methods on the Transformer and Conformer with different encoder layers and hidden dimensions.
We use Stack-4 as the baseline model (see Table \ref{PDS_setting} for the setting).
For a fair comparison, the number of model parameters in each group is similar.

For a popular setting of 12 encoder layers with 256 hidden dimensions in group (A), PDS achieves a very high down-sampling ratio of 32 with a slight improvement.
As a bonus, it yields a speedup of 1.20$\times$.
In real scenarios, the saved computational resource can be reinvested in a bigger batch or improved model capacity.
As the down-sampling ratio decreases, the performance improves significantly.
Similar phenomena are observed on the wider Transformer with 512 hidden dimensions in group (B), where our method benefits more in terms of speedup.

Interestingly, we find that the deep models with 30 encoder layers in group (C) eliminate the performance gap under different down-sampling ratios.
PDS compresses the representation to 1/32 of the initial length while achieving a considerable relative reduction of 0.73 WER points.
We conjecture that the deep model allows more sufficient modeling in each stage and preserves the information even in an extreme case of down-sampling.
This also has practical advantages in industrial scenarios where deep models are preferred.

We observe two interesting phenomena in the experiments on Conformer architecture in groups (D) and (E).
Firstly, Conformer bridges the performance gap between the stack and PDS method.
We speculate that the Conformer integrates the relative position encoding (RPE) to improve the generalization to the variant length, which may be helpful for the long sequence encoding and reduces the benefit of our method.
Secondly, PDS with a ratio of 16 outperforms its counterpart with a lower ratio of 8.
Conformer enhances local interaction among neighbor contexts by convolution module, which leads to a higher similarity, as shown in Figure \ref{pds_similarity}.

We also notice that PDS works better on clean subsets than other subsets, especially under a high down-sampling ratio. 
This demonstrates that down-sampling is more challenging on noisy audio, where it is difficult to distinguish the meaningful information. 
We will explore more robust methods in future work.


\begin{table}[t!]
  \centering
  \setlength{\tabcolsep}{3mm}{
  \footnotesize
  \begin{tabular}{l|c|c|c|c}
    \toprule
    \multirow{2}*{Setting} & \multicolumn{2}{c|}{w/o CTC} & \multicolumn{2}{c}{w/ CTC} \\
    \cmidrule(lr){2-5}
    & dev & test & dev & test  \\
    \midrule
    Stack-4 & 5.42 & 5.80 & 4.96 & 5.55 \\
    \specialrule{0em}{1pt}{1pt}
    \cdashline{1-5}
    \specialrule{0em}{1pt}{1pt}
    PDS-Base-8 & \textbf{5.09} & \textbf{5.59} & \textbf{4.72} & \textbf{5.24} \\
    PDS-Base-16 & 5.30 & 5.78 & 4.76 & 5.28 \\
    PDS-Base-32 & 5.43 & 5.85 & 5.17 & 5.63 \\
    \bottomrule
  \end{tabular}
  }
  \caption{WER on the AISHELL-1 ASR corpus.}
  \label{res_aishell}
\end{table}

\begin{table}[t!]
  \centering
  \footnotesize
  \begin{tabular}{l|c|r|c|c}
    \toprule
    {\footnotesize Setting} & {\footnotesize CTC} & {\footnotesize \#Params} & {\footnotesize w/o PT} & {\footnotesize w/ PT} \\
    \midrule
    \multicolumn{4}{l}{Transformer} \\
    \specialrule{0em}{1pt}{1pt}
    \cline{1-5}
    \specialrule{0em}{1pt}{1pt}
    \multirow{2}*{Stack-4} &  & 30M & 20.2 & 23.2 \\
     & \checkmark             & 32M & 24.0 & \textbf{24.5} \\
    \specialrule{0em}{1pt}{1pt}
    \cdashline{1-5}
    \specialrule{0em}{1pt}{1pt}
    \multirow{2}*{PDS-Base-8} & & 29M & 23.2 & 24.5 \\
     & \checkmark & 32M & 24.2 & \textbf{24.8} \\
    \midrule
    \multicolumn{4}{l}{SATE - Transformer} \\
    \specialrule{0em}{1pt}{1pt}
    \cline{1-5}
    \specialrule{0em}{1pt}{1pt}
    Stack-4    & \checkmark & 40M & 24.8 & 25.3 \\
    PDS-Base-8 & \checkmark & 40M & 25.5 & \textbf{25.6} \\
    \midrule
    \multicolumn{4}{l}{SATE - Conformer} \\
    \specialrule{0em}{1pt}{1pt}
    \cline{1-5}
    \specialrule{0em}{1pt}{1pt}
    Stack-4    & \checkmark & 55M & 25.5 & 25.9  \\
    PDS-Base-8 & \checkmark & 55M & 25.8 & \textbf{26.4} \\
    \midrule
    \multicolumn{4}{l}{SATE - Conformer - Unrestricted} \\
    \specialrule{0em}{1pt}{1pt}
    \cline{1-5}
    \specialrule{0em}{1pt}{1pt}
    Stack-4    & \checkmark & 130M & - & 27.9  \\
    PDS-Base-8 & \checkmark & 134M & - & \textbf{28.7} \\
    \bottomrule
  \end{tabular}
  \caption{SacreBLEU on the MuST-C En-De ST corpus. "PT" represents the ST models are initialized with the pre-trained ASR and MT models.
  }
  \label{res_st}
\end{table}

\noindent \textbf{AISHELL-1} We observe that our method with a down-sampling ratio of 32 slightly under-performs the baseline on the AISHELL-1 corpus, as shown in Table \ref{res_aishell}.
As we employ a character vocabulary, the length of the compressed sequence may be less than transcription, making excessive compression and invalid connectionist temporal classification (CTC) computation.
This inspires us to explore better solutions, e.g., a self-adaptive compression method that dynamically treats each sample.
We further discuss it in Section Limitations.

\subsection{Results of ST}

End-to-end ST has become popular recently \cite{Berard_arxiv2016}.
However, unlike ASR, annotated speech-to-translation data is scarce, making it challenging for well-trained ST models.
Therefore, CTC and pre-training methods are used for sufficient training \cite{Bahar_ASRU2019, zhang2022improving}.
According to the results of ASR, we select PDS-Base-8 to investigate the effects of PDS on both performance and model convergence.

Table \ref{res_st} shows a substantial performance gap of 3.0 BLEU points between the stack and PDS methods when the auxiliary CTC and pre-training methods are not used.
This indicates that bridging the granularity gap helps convergence and improves ST when transcription is not available.
With CTC and pre-training, better performance is achieved by strong supervision and better initialization.
Also, PDS outperforms the stack method significantly.

Better architecture of SATE \cite{Xu_ACL2021} brings consistent improvements.
The encoder of SATE is composed of an acoustic encoder and a textual encoder.
We only employ PDS in the acoustic encoder.
Although an adaptor is introduced for adaptive representation, the length inconsistency issue is not solved in the original implementation.
As a popular method, the shrink mechanism filters the acoustic representation based on the CTC prediction \cite{Dong_aaai2021}.
However, it also poses the risk of information loss due to inaccurate predictions.
PDS provides another approach by generating a length-matched sequence in foundational acoustic encoding.

Combing the CTC and pre-training methods, PDS achieves a competitive performance of 26.4 BLEU scores without additional training data. 
Under the more challenging unrestricted setting, PDS provides a better acoustic representation and yields a remarkable improvement of 0.8 BLEU points over the stack method.

\section{Analysis}

\begin{table}[tbp]
  \centering
  \footnotesize
  \begin{tabular}{c|c|r|l|l|c}
    \toprule
    Group & {\footnotesize F} & 
    {\footnotesize Ratio}  & 
    {\footnotesize Stride} & 
    {\footnotesize  Layer} & 
    {\footnotesize Avg.}   \\
    \midrule
    \multicolumn{4}{l}{Stack} \\
    \specialrule{0em}{1pt}{1pt}
    \cline{1-6}
    \specialrule{0em}{1pt}{1pt}
    \multirow{4}{*}{(A)} & \multirow{4}{*}{/} & 2 & 2 & 12 & 7.06 \\
    & & 4 & 2-2 & 0-12 & \textbf{6.80} \\
    & & 8 & 2-2-2 & 0-0-12 & 7.43 \\
    & & 16 & 2-2-2-2 & 0-0-0-12 & 9.17 \\
    \midrule
    \multicolumn{4}{l}{PDS} \\
    \specialrule{0em}{1pt}{1pt}
    \cline{1-6}
    \specialrule{0em}{1pt}{1pt}
    \multirow{2}{*}{(B)} &  & 4 & 2-2-1-1 & 3-3-3-3 & 6.01 \\
    & \checkmark & 4 & 2-2-1-1 & 3-3-3-3 & \textbf{5.99} \\
    \specialrule{0em}{1pt}{1pt}
    \cline{1-6}
    \specialrule{0em}{1pt}{1pt}
    \multirow{2}{*}{(C)} &  & 8 & 2-2-1-2 & 3-3-3-3 & 6.53 \\
    & \checkmark & 8 & 2-2-1-2 & 3-3-3-3 & \textbf{6.11} \\
    \specialrule{0em}{1pt}{1pt}
    \cline{1-6}
    \specialrule{0em}{1pt}{1pt}
    \multirow{4}{*}{(D)} &  & 8 & 2-2-2-1 & 2-2-6-2 & 6.55 \\
    & & 16 & 2-2-2-2 & 2-2-6-2 & 6.77 \\
    & \checkmark & 8 & 2-2-2-1 & 2-2-6-2 & 6.28 \\
    & \checkmark & 16 & 2-2-2-2 & 2-2-6-2 & \textbf{6.26} \\
    \specialrule{0em}{1pt}{1pt}
    \cline{1-6}
    \specialrule{0em}{1pt}{1pt}
    \multirow{2}{*}{(E)} &  & 32 & 2-2-2-2-2 & 2-2-3-3-2 & 7.21 \\
    & \checkmark & 32 & 2-2-2-2-2 & 2-2-3-3-2 & \textbf{6.69} \\
    \bottomrule
  \end{tabular}
  \caption{Impact of representation fusion. "F" represents the representation fusion method. We report the average WER of all 4 sets of LibriSpeech.}
  \label{why_fusion}
\end{table}

Next, we study a number of interesting problems on LibriSpeech. We present the comparisons with previous work and more analyses in Appendix \ref{add_analyses}.

\subsection{Impact of Representation Fusion}
\label{analysis_drop}

To investigate the impact of information loss and the importance of representation fusion, we compare the results under different down-sampling ratios (see Table \ref{why_fusion}).

The standard setting for the stack method is to down-sample the input with a lower ratio of 4.
This setting also achieves the best performance in group (A).
A lower ratio of 2 leads to inferior WER because long but fine-grained features face the modeling challenges.
As the ratio of down-sampling increases, the performance drops drastically.
This supports the point that information loss is severe in the stack method.

The PDS method outperforms the stack method under the same setting of ratio $=4$.
However, we find that the fusion method does not obtain significant improvement.
This may be because the lightweight compression is lossless and cannot benefit from the fusion of representations.

Interestingly, the fusion method achieves consistent improvements when higher down-sampling ratios are employed.
To study it further, we design another set of experiments with a special setting in group (D): the down-sampling ratio decreases from 16 to 8 by setting the stride of the final stage to 1.
Then, we achieve a better performance of 6.55 WER points, which indicates less information loss under a slighter compression.
With the help of the fusion method, two settings achieve similar performances.

\begin{table}[tbp]
  \centering
  \footnotesize
  \begin{tabular}{l|c|c|c|c|c}
    \toprule
    \multirow{2}*{Layer} & \multicolumn{2}{c|}{{\footnotesize dev}} & \multicolumn{2}{c|}{{\footnotesize test}} & \multirow{2}*{Avg.} \\
    \cmidrule(lr){2-3} \cmidrule(lr){4-5}
    & {\footnotesize clean} & {\footnotesize other} & {\footnotesize {\footnotesize clean}} & {\footnotesize other} & \\
    \midrule
    2-2-2-6 & 3.91 & 9.29 & 4.09 & 9.38 & 6.63 \\
    2-2-4-4 & 3.83 & 9.11 & 4.05 & 9.13 & 6.49 \\  
    2-2-6-2 & 3.71 & 8.73 & 3.74 & 9.02 & \textbf{6.26} \\
    \midrule
    5-5-10-10 & 3.19 & 7.79 & 3.57 & 7.69 & 5.52 \\
    5-5-12-8  & 3.15 & 7.83 & 3.38 & 7.79 & \textbf{5.50} \\
    5-5-15-5  & 3.27 & 7.59 & 3.60 & 7.83 & 5.53 \\
    \bottomrule
  \end{tabular}
  \caption{Impact of the number of layers in each stage. We report the results of Transformer with PDS-Base-16 and PDS-Deep-16 settings.}
  \label{layers}
\end{table}

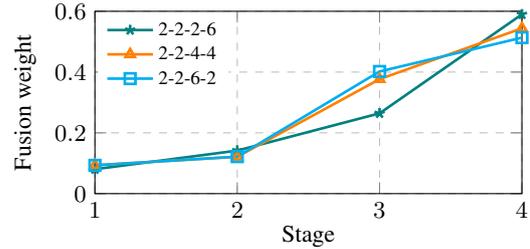
\begin{figure}[tbp]
  \centering
    \begin{tikzpicture}
      \footnotesize{
      \begin{axis}[
        ymajorgrids,
        xmajorgrids,
        grid style=dashed,
        width=.45\textwidth,
        height=0.25\textwidth,
        legend columns=1,
        legend entries={2-2-2-6, 2-2-4-4, 2-2-6-2},
        legend style={
          draw=none,
          line width=1pt,
        },
        legend style={at={(0.16,1.0)}, anchor=north,
        nodes={scale=0.8, transform shape}},
        xmin=1, xmax=4,
        ymin=0, ymax=0.6,
        xtick={1, 2, 3, 4},
        xlabel=\footnotesize{Stage},
        ylabel=\footnotesize{Fusion weight},
        ylabel style={yshift=-1em},
        xlabel style={yshift=0.7em},
        scaled ticks=false,
        ]
    \addplot[teal, mark=star, line width=1pt] 
    coordinates{
        (1, 0.0804) (2, 0.1416) (3, 0.2642) (4, 0.5898)
    };
    \addplot[orange, mark=triangle, line width=1pt] 
    coordinates{
        (1, 0.0936) (2, 0.1212) (3, 0.3767) (4, 0.5439)
    };
    \addplot[cyan, mark=square, line width=1pt] 
    coordinates{
        (1, 0.0931) (2, 0.1218) (3, 0.4019) (4, 0.5132)
    };
    \end{axis}
    }
  \end{tikzpicture}
  \caption{The fusion weights of the output representation in each stage. We consider three settings of the number of layers under PDS-Base-16.}
  \label{weight}
\end{figure}

\begin{figure*}[tbp]
  \begin{tikzpicture}
    \footnotesize{
      \begin{axis}[
        fill opacity=1,
        fill=orange,
        ymajorgrids,
        xmajorgrids,
        grid style=dashed,
        width=.35\textwidth,
        height=0.25\textwidth,
        legend columns=-1,
        legend entries={PDS-Base-8, PDS-Base-16, PDS-Base-32},
        legend style={fill opacity=0.5,text opacity =1,
          draw=none,
          line width=1pt,
        },
        legend style={at={(1.1,1.2)}, anchor=north,
        nodes={scale=0.75, transform shape}},
        xmin=0, xmax=12,
        ymin=20, ymax=100,
        xtick={0,2,...,12},
        xlabel=\footnotesize{Layer},
        ylabel=\footnotesize{Similarity (\%)},
        ylabel style={yshift=-2em},
        xlabel style={yshift=0.7em},
        scaled ticks=false,
        ]
    \addplot[cyan, mark=*, line width=1pt] coordinates{(0,0)};
    \addplot[teal, mark=square, line width=1pt] coordinates{(0,0)};
    \addplot[orange, mark=diamond, line width=1pt] coordinates{(0,0)};
    \addplot[cyan, mark=*, line width=1pt,mark indices={1,4,10}] 
    file {data/Transformer-PDS-8.txt};
    \addplot[teal, mark=square, line width=1pt,mark indices={1,3,5,11}] 
    file {data/Transformer-PDS-16.txt};
    \addplot[orange, mark=diamond, line width=1pt,mark indices={1,3,5,8,11}] 
    file {data/Transformer-PDS-32.txt};
    \end{axis}
    }

  \node[align=left] (a) at (1.9,-1.0) {(a) Transformer};
  \hspace{5.3cm}
  \footnotesize{
    \begin{axis}[
      ymajorgrids,
      xmajorgrids,
      grid style=dashed,
      width=.35\textwidth,
      height=0.25\textwidth,
      legend columns=1,
      legend style={fill opacity=0.5,text opacity =1,
        draw=none,
        line width=1pt,
      },
      legend style={at={(0.7,0.45)}, anchor=north,
      nodes={scale=0.75, transform shape}},
      xmin=0, xmax=12,
      ymin=20, ymax=100,
      xtick={0,2,...,12},
      xlabel=\footnotesize{Layer},
      ylabel=\footnotesize{Similarity (\%)},
      ylabel style={yshift=-2em},
      xlabel style={yshift=0.7em},
      scaled ticks=false,
      ]
  \addplot[cyan, mark=*,     line width=1pt,mark indices={1,4,10}] 
  file {data/Conformer-PDS-8.txt};
  \addplot[teal, mark=square, line width=1pt,mark indices={1,3,5,11}] 
  file {data/Conformer-PDS-16.txt};
  \addplot[orange, mark=diamond, line width=1pt,mark indices={1,3,5,8,11}] 
  file {data/Conformer-PDS-32.txt};
  \end{axis}
  }
  \node[align=left] (b) at (1.9,-1.0) {(b) Conformer};
  \hspace{5.3cm}
  \footnotesize{
    \begin{axis}[
      ymajorgrids,
      xmajorgrids,
      grid style=dashed,
      width=.35\textwidth,
      height=0.25\textwidth,
      legend columns=-1,
      legend entries={PDS-Deep-8, PDS-Deep-16, PDS-Deep-32},
      legend style={fill opacity=0.5,text opacity =1,
        draw=none,
        line width=1pt,
      },
      legend style={at={(0.4,1.2)}, anchor=north,
      nodes={scale=0.65, transform shape}},
      xmin=0, xmax=30,
      ymin=20, ymax=100,
      xlabel=\footnotesize{Layer},
      ylabel=\footnotesize{Similarity (\%)},
      ylabel style={yshift=-2em},
      xlabel style={yshift=0.7em},
      scaled ticks=false,
      ]
  \addplot[myblue, mark=*, line width=1pt] coordinates{(0,0)};
  \addplot[ugreen!120, mark=square, line width=1pt] coordinates{(0,0)};
  \addplot[myred!80, mark=diamond, line width=1pt] coordinates{(0,0)};
  \addplot[myblue, mark=*, line width=1pt,mark indices={1,8,22}] 
  file {data/Trasnformer-Deep-PDS-8.txt};
  \addplot[ugreen!120, mark=square, line width=1pt,mark indices={1,6,11,23}] 
  file {data/Trasnformer-Deep-PDS-16.txt};
  \addplot[myred!80, mark=diamond, line width=1pt,mark indices={1,6,11,18,25},] 
  file {data/Trasnformer-Deep-PDS-32.txt};
  \end{axis}
  }
  \node[align=left] (c) at (1.9,-1.0) {(c) Deep Transformer};
\end{tikzpicture}
  \caption{Similarities of window size of 2 in each layer of Transformer, Conformer, and deep Transformer. The marked points represent the similarity before each down-sampling.}
  \label{pds_similarity}
\end{figure*}
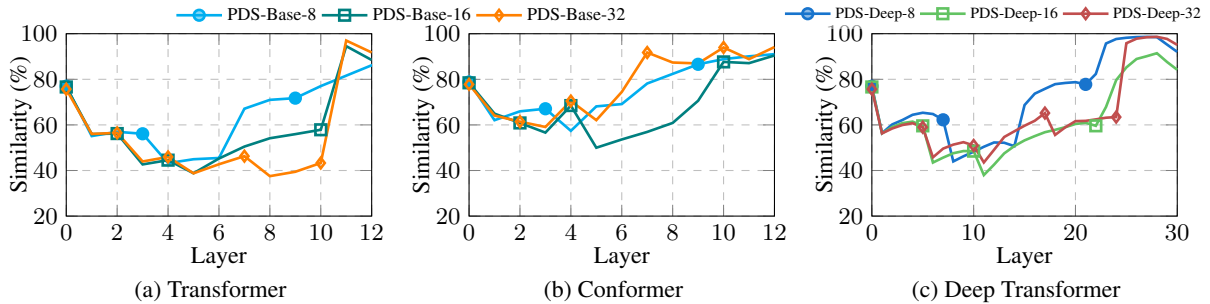

\subsection{Impact of Model Depth}

We compare the performance of the different number of layers in each stage.
Table \ref{layers} shows the results on base and deep models.

For the model of 12 encoder layers, we assign 2 layers in the bottom 2 stages for less computation cost.
As described in Section \ref{details}, PDS achieves better performance as the number of layers increases in the intermediate stages.
There are two reasons.
Firstly, the information loss in the intermediate stages is less compared with the top stage, and thus the increased encoding layers are helpful.
Secondly, sufficient encoding provides high similarity and helps the down-sampling for the next stage, but it is not the case for the top stage.
This is consistent with the previous conclusion \cite{Huang_Interspeech2020}.
We also show the fusion weights in Figure \ref{weight}.
The weight increases as the number of layers increases, and vice versa.

Furthermore, we compare the results of deep Transformer models with a 30-layer encoder.
Due to the sufficient encoding in each stage, the deep model is robust in the design of the number of layers.
There is no obvious performance gap across the three different settings.
It is very meaningful to combine PDS with the popular deep models \cite{Pham_Interspeech2019} in the follow-up work.

\subsection{Impact on Similarity}
\label{impact_on_corr}

Unlike the stack method, PDS performs the context interaction process after each down-sampling process.
Figure \ref{pds_similarity} shows the similarity across different model architectures.

In Transformer, high similarity (about $60\% \sim 80\%$) alleviates the information loss under the setting of PDS-Base-8.
As the down-sampling ratio increases, fewer layers in each stage cannot capture the context information sufficiently and thus make the degraded similarity and worse performance.

Despite the limited layers in each stage, Conformer always shows high similarities due to explicit local modeling.
This also demonstrates the effectiveness of Conformer.
The deep Transformer alleviates the issue directly by stacking more encoder layers.

One interesting finding is that the similarity of top layers is very high ($> 90\%$) across all architectures, which may be due to the effect of the direct supervision from the decoder.
This inspires us to explore multi-task learning methods by injecting explicit training objectives into intermediate stages.

\subsection{Distribution of Attention Weights}

\begin{figure}[tbp]
  \centering
    \begin{tikzpicture}
      \footnotesize{
      \begin{axis}[
        ymajorgrids,
        xmajorgrids,
        grid style=dashed,
        width=.45\textwidth,
        height=0.3\textwidth,
        legend columns=1,
        legend entries={Stack-4,PDS-Base-8, PDS-Base-16, PDS-Base-32},
        legend style={
          draw=none,
          legend cell align=left,
          line width=1pt,
        },
        ycomb,
        area style,
        legend style={at={(0.76,1.0)}, anchor=north,
        nodes={scale=0.8, transform shape}},
        xmin=0, xmax=2,
        ymin=0, ymax=8,
        xlabel=\footnotesize{Attention weight},
        ylabel=\footnotesize{Percentage (\%)},
        ylabel style={yshift=-1.5em},
        xlabel style={yshift=0.5em},
        scaled ticks=false,
        yticklabel style={/pgf/number format/precision=2,/pgf/number format/fixed},
        ]
    \addplot[myred!90,fill=myred!90, opacity=0.8,line width=0.7pt] 
    file {data/Distribution-base.txt};
    \addplot[teal!85,fill=teal!85, opacity=0.8,line width=0.7pt] 
    file {data/Distribution-PDS-8.txt};
    \addplot[ugreen!85,fill=ugreen!85,opacity=0.8,line width=0.7pt] 
    file {data/Distribution-PDS-16.txt};
    \addplot[orange!65,fill=orange!65,opacity=0.8,line width=0.7pt] 
    file {data/Distribution-PDS-32.txt};    
    \end{axis}
    }
  \end{tikzpicture}
  \caption{Distribution of summed cross-attention weights for each encoder representation on LibriSpeech test-clean set.}
  \label{attention_distribution}
\end{figure}
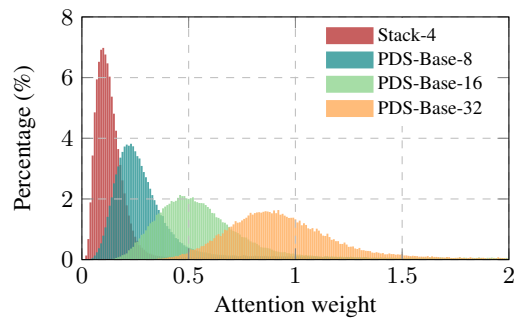

Our method bridges the granularity gap and generates semantically more complete units, e.g., text-level representations.
We suppose that this informative representation has a greater effect on text generation.
Referring to \newcite{Zhang_EMNLP2020}, Figure \ref{attention_distribution} shows the distribution of summed cross-attention weights for each encoder representation.

The fine-grained representations in the stack method have a remarkable granularity gap with the text representations.
Therefore, the smaller attention weights must spread across multiple relevant representations in cross-attention operation, which makes it hard to capture complete information for text generation.
In our method, each representation receives greater weights as the down-sampling ratio increases.
This indicates that each unit has a more meaningful contribution to the generation, and it is easier to capture the relevant source information.

\section{Conclusion}

In this paper, we explore how to compress the fine-grained frame-level acoustic features to the semantically complete text-level units, like character-level and subword-level representations.
To alleviate this granularity gap, we first investigate the down-sampling process and reveal the risk of information loss in the popular stack method.
This inspires us to propose \textit{Progressive Down-Sampling}, which gradually attains the coarser-grained representation during encoding.
Furthermore, we develop a representation fusion method to combine the high-level and low-level information, which is important for high down-sampling ratios.
By this, we are the first to compress the acoustic features with a ratio of 32 on the ASR task while achieving comparable or even better performances.
The more challenging ST task demonstrates that the alleviated granularity gap facilitates convergence effectively.

\section{Limitations}

Many challenges remain in the follow-up of our work.
Here are some limitations that we intend to resolve in the future:
\begin{itemize}
    \item We need a more robust method for compression. Although our method achieves consistent improvements in most experiments, we notice that the benefit is limited in the noisy sets, especially under a high down-sampling ratio. This drives us to develop a more robust down-sampling method for preserving meaningful information even with high compression.
    \item Our method compresses all the input acoustic features with the same ratio, where the ratio is determined according to the whole dataset. However, the speed of each audio is different, which results in obstacles to unified down-sampling. Ideally, each sample should be compressed with a self-adaptive ratio. 
\end{itemize}

\section*{Acknowledgement}

The authors would like to thank anonymous reviewers for their insightful comments.
This work was supported in part by the National Science Foundation of China (No. 62276056), the National Key R\&D Program of China, the China HTRD Center Project (No. 2020AAA0107904), the Natural Science Foundation of Liaoning Province of China (2022-KF-16-01), the Yunnan Provincial Major Science and Technology Special Plan Projects (No. 202103AA080015), the Fundamental Research Funds for the Central Universities (Nos. N2216016, N2216001, and N2216002), and the Program of Introducing Talents of Discipline to Universities, Plan 111 (No. B16009).

\bibliography{anthology,custom}
\bibliographystyle{acl_natbib}

\appendix

\label{sec:appendix}

\section{Experimental Details}
\label{appendix_exp}

\subsection{Datasets and Pre-processing}

The datasets are from three benchmarks:

\begin{itemize}
\item {\textbf{LibriSpeech}} is a publicly available read English ASR corpus, which consists of 960-hour training data \cite{Panayotov_ICASSP2015}. 
The development and test data are divided into clean and other subsets according to the speech quality. 
We select the model on the dev-clean set and report results on all four subsets, including dev-clean, dev-other, test-clean, and test-other.
The average WER is computed on the concatenation of all four subsets.

\item {\textbf{AISHELL-1}} is a publicly available Chinese Mandarin speech corpus, which consists of 170-hour training data \cite{Hu_aishell}.
We select the model on the dev set and report results WER on the dev and test sets.

\item {\textbf{MuST-C}} is a multilingual speech translation corpus extracted from the TED talks \cite{Gangi_NAACL2019}. We train our systems on the English-German speech translation dataset of 400-hour speech. We select (and tune) the model on the dev set and report results on the tst-COMMON set.
\end{itemize}

For pre-processing, we follow the common recipes in fairseq toolkit\footnote{\href{https://github.com/pytorch/fairseq}{https://github.com/pytorch/fairseq}}, which removes the utterances of more than 3,000 frames or fewer than 5 frames. 
We only use the speed perturbation in the experiments built on the AISHELL-1.
The 80-channel Mel filter bank features are extracted by a 25ms window with a stride of 10ms.
We learn SentencePiece\footnote{\href{https://github.com/google/sentencepiece}{https://github.com/google/sentencepiece}} segmentation with a size of 10,000 for the LibriSpeech and MuST-C datasets, and use 4231 characters for the AISHELL-1 dataset.
For the MuST-C ST dataset, we use a shared vocabulary for the source and target languages.

\begin{table*}[thbp]
  \centering
  \footnotesize
  \begin{tabular}{c|c|c|c|c|c|c|r|c|r|c}
    \toprule
    \multirow{2}*{Group} &
    \multirow{2}*{\footnotesize Arch} & 
    \multirow{2}*{\footnotesize Ratio} & 
    \multirow{2}*{\footnotesize Layer} & 
    \multirow{2}*{\footnotesize  Hidden Size} & 
    \multirow{2}*{\footnotesize  \#Params.} & 
    \multicolumn{2}{c|}{{\footnotesize dev}} & 
    \multicolumn{2}{c|}{{\footnotesize test}} &
    \multirow{2}*{\footnotesize Avg.}   \\
    \cmidrule(lr){7-8} \cmidrule(lr){9-10}
    & & & & & & {\footnotesize clean} & {\footnotesize other} & {\footnotesize clean} & {\footnotesize other} \\
    \midrule
    \multicolumn{4}{l}{Stack-4} \\
    \specialrule{0em}{1pt}{1pt}
    \cline{1-11}
    \specialrule{0em}{1pt}{1pt}
    
    \multirow{2}{*}{(A)} 
    & CTC &\multirow{2}{*}{4}& \multirow{2}{*}{12}& \multirow{2}{*}{256}& 20M & 5.63 & 13.53 & 5.73 & 13.39 & 9.50 \\& Enc-Dec & & & & 30M & 3.88 & 9.26 & 4.49 & 9.42 & \textbf{6.80} \\
    \midrule
    \multicolumn{4}{l}{PDS-8} \\
    \specialrule{0em}{1pt}{1pt}
    \cline{1-11}
    \specialrule{0em}{1pt}{1pt}
    \multirow{3}{*}{(B)} & 
    \multirow{3}{*}{CTC} &
    \multirow{3}{*}{2-2-1-2}& 3-3-3-3 & 256-256-256-256 & \multirow{3}{*}{20M}& 5.06 & 11.97 & 5.26 & 11.90 & 8.49  \\
    & & & 3-3-3-3 & 192-256-256-320 & & 5.16 & 12.12	& 5.29 & 12.00 & 8.58 \\
    & & & 5-3-3-5 & 192-224-224-256 & & 4.97 & 11.89 & 5.14 & 11.83 & \textbf{8.40}\\
    \cline{1-11}
    \specialrule{0em}{1pt}{1pt}
    \multirow{3}{*}{(C)} & \multirow{3}{*}{Enc-Dec} & \multirow{3}{*}{2-2-1-2} & 3-3-3-3 & 256-256-256-256 &\multirow{3}{*}{30M} & 3.57 & 8.63 & 3.85 & 8.58 & 6.11 \\
    & & & 3-3-3-3 & 192-256-256-320 &     & 3.43 & 8.38 & 3.71 & 8.48 & \textbf{5.95} \\
    & & & 5-3-3-5 & 192-224-224-256 &     & 3.43 & 8.30 & 3.97 & 8.60 & 6.03 \\
    \midrule
    \multicolumn{4}{l}{PDS-16} \\
    \specialrule{0em}{1pt}{1pt}
    \cline{1-11}
    \specialrule{0em}{1pt}{1pt}
    \multirow{3}{*}{(D)} & 
    \multirow{3}{*}{CTC} & 
    \multirow{3}{*}{2-2-2-2} & 2-2-6-2 & 256-256-256-256 & \multirow{3}{*}{20M} & 5.66 & 12.75 & 5.89 & 12.72 & 9.19 \\
    & & & 2-2-6-2 & 192-224-256-320 & & 5.56 & 12.50 & 5.69	& 12.77 & 9.07 \\
    & & & 3-3-9-3 & 160-192-224-256 & & 5.23 & 11.80 & 5.27 & 11.81 & \textbf{8.47} \\
    \cline{1-11}
    \specialrule{0em}{1pt}{1pt}
    \multirow{3}{*}{(E)} & \multirow{3}{*}{Enc-Dec} & \multirow{3}{*}{2-2-2-2} & 2-2-6-2 & 256-256-256-256	&\multirow{3}{*}{30M}& 3.71 & 8.73 & 3.74 & 9.02 & \textbf{6.26} \\
    & & & 2-2-6-2 & 192-224-256-320 & & 3.92 & 9.18 & 4.13	& 9.00 & 6.51 \\
    & & & 3-3-9-3 & 160-192-224-256 & & 3.79 & 8.78 & 4.19 & 8.83 & 6.35 \\
    \bottomrule
  \end{tabular}
  \caption{Comparison of the settings of hidden dimension. Attention heads are set to 4 and the feed-forward dimension is 4 times the hidden dimension.}
  \label{comparion_dim}
\end{table*}

\subsection{Model Settings}

We use the encoder-decoder framework and implement the method based on the fairseq toolkit.
We use the Adam optimizer and adopt the default learning schedule in fairseq. 
We apply dropout with a rate of 0.1 and label smoothing $\epsilon_{ls} = 0.1$ for regularization.
SpecAugment \cite{Park_ISCA2019} is applied in the input speech features for better generalization and robustness.

For the LibriSpeech ASR task, we evaluate our method on Transformer \cite{Vaswani_nips2017} and Conformer \cite{Gulati_ISCA2020}.
The settings of the encoder for ASR models are shown in Table \ref{res_asr}.
The decoder consists of 6 Transformer layers and the settings are the same as the encoder.
CTC \cite{Graves_ACL2006} multi-task learning is not used due to the very modest improvement in our preliminary experiments.

For the AISHELL-1 ASR task, we evaluate our method on Transformer \cite{Vaswani_nips2017}.
The encoder consists of 12 layers and the decoder consists of 6 layers.
Each layer comprises 256 hidden units, 4 attention heads, and 2,048 feed-forward hidden units.
The weight for CTC multi-task learning is set to 0.3.

For the ST task, we evaluate our method on Transformer and SATE \cite{Xu_ACL2021}. 
Except that the knowledge distillation method is not used for simplicity, we follow the settings of SATE.
The encoder consists of 12 layers for Transformer.
SATE has an acoustic encoder of 12 layers and a textual encoder of 6 layers.
Each layer comprises 256 hidden units, 4 attention heads, and 2,048 feed-forward hidden units.
CTC is employed with a weight of 0.3 for better convergence.
Similar to \newcite{Xu_ACL2021}, we also consider both restricted and unrestricted scenarios.
Under the restricted setting, the ASR and MT models are pre-trained with the MuST-C En-De data.
Under the unrestricted setting, we use the additional LibriSpeech ASR corpus and Opensubtitle En-De MT corpus for pre-training.
We also use Conformer as the acoustic encoder and widen the model by increasing the hidden size to 512 and attention heads to 8.

All the models are trained for 100 epochs.
We early stop training when there is no performance improvement on the development set for 10 consecutive checkpoints.
We use beam search decoding with a beam size of 5 for all models on 1 NVIDIA TITAN RTX GPU.
The CTC and language model re-scoring methods are not used.
We report WER and case-sensitive SacreBLEU for ASR and ST tasks, respectively.

\begin{table*}[h!]
  \centering
  \footnotesize
  \begin{tabular}{c|c|c|c|c|c|c|c|c|c|c}
    \toprule
    \multirow{2}*{\footnotesize Model} & 
    \multirow{2}*{\footnotesize Ratio} & 
    \multirow{2}*{\footnotesize Layer} & 
    \multirow{2}*{\footnotesize  Hidden Size} & 
    \multirow{2}*{\footnotesize  \#Params.} & 
    \multicolumn{2}{c|}{{\footnotesize dev}} & 
    \multicolumn{2}{c|}{{\footnotesize test}} &
    \multirow{2}*{\footnotesize Avg.}  &
    \multirow{2}*{\footnotesize Speedup}   
    \\
    \cline{6-9}
    & & & & & {\footnotesize clean} & {\footnotesize other} & {\footnotesize clean} & {\footnotesize other} &  \\
    \cline{1-11}
    \specialrule{0em}{1pt}{1pt}
    \multicolumn{1}{l|}{Conformer$^\dag$} &\multirow{2}{*}{4} & \multirow{2}{*}{16} & \multirow{2}{*}{174} & \multirow{2}{*}{13.0M} & 5.31 & 13.60 & 5.41 & 13.24 & 9.33 & - \\
    \multicolumn{1}{l|}{Conformer } & & & & & 4.76 & 12.07 & 4.72 & 12.19 & \textbf{8.36} & 1.00$\times$ \\
    \specialrule{0em}{1pt}{1pt}
    \cline{1-11}
    \specialrule{0em}{1pt}{1pt}
    \multicolumn{1}{l|}{{Eff Conformer$^\dag$}} & \multirow{2}{*}{2-2-2} & \multirow{2}{*}{5-5-5} & \multirow{2}{*}{120-168-240} & \multirow{2}{*}{13.3M} & 4.46 & 11.36 & 4.61 & 11.29 & 7.88 & - \\
    \multicolumn{1}{l|}{{Eff Conformer}} & & & & & 4.25 & 11.14 & 4.35 & 10.99 & \textbf{7.56} & 1.12$\times$ \\
    \specialrule{0em}{1pt}{1pt}
    \cline{1-11}
    \specialrule{0em}{1pt}{1pt}
    \multicolumn{1}{l|}{PDS} & \multirow{3}{*}{2-2-2} & \multirow{3}{*}{5-5-5} & 174-174-174 & 13.0M & 4.35 & 11.66 & 4.39 & 11.61 & 7.93 & 1.26$\times$ \\
    \multicolumn{1}{l|}{PDS} & & & 120-168-240 & 12.8M & 4.04 & 10.87 & 4.16 & 10.71 & \textbf{7.37} & 1.34$\times$ \\
    \multicolumn{1}{l|}{\ \ -Fusion} & & & 120-168-240 & 12.5M & 4.29 & 11.02 & 4.29 & 11.18 & 7.65  & 1.36$\times$ \\
    \specialrule{0em}{1pt}{1pt}
    \bottomrule
  \end{tabular}
  \caption{Comparison with effective Conformer \cite{Burchi_Corr2021}. Note that $\dag$ represents that the results are reproduced by released code. We run other experiments based on our codebase.}
  \label{comparison_with_effe}
\end{table*}

\section{Additional Analyses}
\label{add_analyses}
\subsection{Comparison of Dimension Settings}
\label{appendix_compare_dim}

Researchers explore similar pyramid architectures in the field of computer vision \cite{He_CVPR2016, Wang_Corr2021} and acoustic encoding \cite{Burchi_Corr2021}.
A typical setting is that the hidden dimension increases from bottom to top, while the sequence length decreases accordingly.
Although this is reasonable to keep the same complexity, the detailed design needs more tuning efforts.

We make a preliminary exploration of the dimension design on the ASR task.
The final output dimension is defined as $d$, and we consider three settings:
\begin{itemize}
    \item \textit{Same}. This is the basic setting, and the hidden dimensions across the whole encoder are $d$.
    \item \textit{Width-growth}. The hidden dimensions increase from bottom to top, and the middle dimension is set to $d$. This makes a wider model on the topmost stage under the same model parameters.
    \item \textit{Depth-growth}. The hidden dimensions increase from bottom to top, and the dimension of the topmost stage is set to $d$. This makes a deeper model under the same model parameters.
\end{itemize}

In Table \ref{comparion_dim}, we evaluate results of two popular ASR architectures: CTC-based and encoder-decoder models.
CTC limits that the input length must be longer than the corresponding label, so we do not build the experiments with a down-sampling ratio of 32.
For each group of experiments, we list the results of \textit{same}, \textit{width-growth}, and \textit{depth-growth} in turn.
In the CTC-based model in group (B) and group (D), the \textit{depth-growth} setting is superior under both two down-sampling ratios, where the parameters are more efficient than other settings.
Especially under a high down-sampling ratio of 16, the improvement is more significant due to the more sufficient encoding before down-sampling, as shown in Section \ref{impact_on_corr}.
The \textit{wider-growth} setting only achieves the comparable performance with the \textit{same} setting.
This also demonstrates that the deeper models have more potential than the wider models under the same parameters \cite{Wang_ACL2019}.

However, different results appear in the encoder-decoder models.
As shown in group (C) and group (E), the \textit{width-growth} and \textit{depth-growth} settings can not bring consistent improvements.
We conjecture that the CTC-based model suffers from a heavy encoding burden due to the encoder-only modeling.
Therefore, the parameter-efficient designs of the growth settings improve the model capacity remarkably.
The decoder alleviates the generation burden, which leads to the modest improvement of the growth settings.
There are similar results in \newcite{Burchi_Corr2021}, where the improvements in RNN-T models reduce compared with the CTC-based models.

According to the above results, we use the \textit{same} setting in our experiments for simplification.
We will explore parameter-efficient designs in the future.

\subsection{Comparison with Previous Work}

\newcite{Burchi_Corr2021} propose the efficient Conformer, which implements the progressive down-sampling by the modified convolutional module or the strided self-attention module. 
Compared with the efficient Conformer, our method does not rely on the specific architecture. 
We use a lightweight and pluggable module for down-sampling, which allows flexible integration with other methods.

To demonstrate the effectiveness of our method, we construct the comparison of our method and efficient Conformer on the ASR task.
Following \newcite{Burchi_Corr2021}, we learn SentencePiece segmentation with a size of 256.
The grouped multi-head attention is a general method and is not used for a fair comparison.
Except that the models are trained for 100 epochs rather than 450 epochs for fast comparison, we use the same hyperparameters.

As shown in Table \ref{comparison_with_effe}, our method achieves a reduction of 0.19 WER points compared with efficient Conformer.
We also show the result of PDS with the \textit{same} settings. 
In this more challenging CTC-based Conformer model, the \textit{width-growth} setting is more parameter-efficient.
Without the representation fusion method, we only achieve comparable performance with the efficient Conformer. 
This proves the importance of the fusion method that alleviates information loss by combining multi-level representations.

In terms of speedup, our method encodes with more efficient computation due to the simplicity of design. It also shows that the representation fusion is lightweight but brings significant improvements.
Most efficient attention variants can be integrated into our method, which has enormous potential for fast inference.
But we focus on the design of the basic architecture and leave the exploration to the future.

\begin{figure}[t!]
  \centering
    \begin{tikzpicture}
      \footnotesize{
      \begin{axis}[
        ymajorgrids,
        xmajorgrids,
        grid style=dashed,
        width=.45\textwidth,
        height=0.25\textwidth,
        legend columns=4,
        legend entries={Stack-4, PDS-8, PDS-16, PDS-32},
        legend style={
          draw=none,
          line width=1pt,
        },
        legend style={at={(0.5,1.0)}, anchor=south,
        nodes={scale=0.8, transform shape}},
        xmin=100, xmax=960,
        ymin=3, ymax=18,
        xtick={100, 460, 960},
        xlabel=\footnotesize{Size of training data (Hours)},
        ylabel=\footnotesize{WER},
        ylabel style={yshift=-1.5em},
        xlabel style={yshift=0.0em},
        yticklabel style={/pgf/number format/precision=0,/pgf/number format/fixed zerofill},
        scaled ticks=false,
        ]
    \addplot[myred, mark=star, line width=1pt] 
    coordinates{
        (100, 14.46) (460, 6.53) (960, 4.49)
    };
    \addplot[cyan, mark=*, line width=1pt] 
    coordinates{
        (100, 11.55) (460, 5.43) (960, 3.85)
    };
    \addplot[teal, mark=square, line width=1pt] 
    coordinates{
        (100, 12.29) (460, 5.35) (960, 3.74)
    };
    \addplot[orange, mark=diamond, line width=1pt] 
    coordinates{
        (100, 15.94) (460, 5.4) (960, 4.21)
    };
    \end{axis}
    }
  \end{tikzpicture}
  \caption{Comparison of varying amounts of training data. We report the WER on the test-clean set.}
  \label{low_resource}
\end{figure}
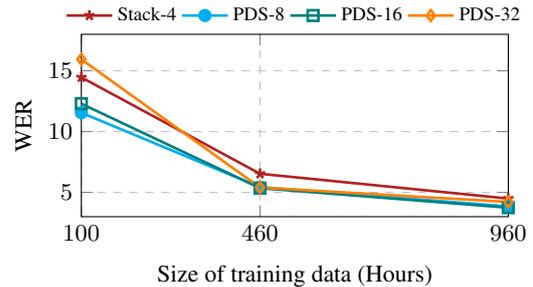

\subsection{Low Resource Setting}

Results on ST show that our method helps convergence significantly.
To further verify it, we compare the results of varying amounts of training data.
We train the base Transformer of 12 encoder layers on the LibriSpeech with common subsets of 100h, 460h, and all 960h training data.
We select the appropriate hyperparameter for each model, including learning rate, batch size, and CTC multi-task learning.
The WER on the test-clean set is reported in Figure \ref{low_resource}.

Under the more challenging setting where only 100 hours are available, the lower down-sampling ratio of 8 yields a remarkable improvement of 2.91 WER.
Our method compresses the representations into coarser-grained units, which eases the burden of attention calculation.
But excessive compression leads to degraded performance due to inferior CTC computation.
One precondition for CTC loss is that the length of the input must be longer than the length of the corresponding label, which leads to invalid CTC learning for some samples.
This is consistent with our previous conclusion: we need to develop a self-adaptive method that down-samples each sample with moderate compression.
Increasing the training data also brings consistent improvements across different compression ratios.




\end{document}